\documentclass[11pt]{article}

\usepackage{EMNLP2022}
\usepackage{soul}
\usepackage{changepage,threeparttable} 
\usepackage{times}
\usepackage{latexsym}

\usepackage[T1]{fontenc}

\usepackage[utf8]{inputenc}

\usepackage{microtype}

\usepackage{graphicx}
\usepackage{booktabs}
\usepackage{caption}
\usepackage{subcaption}
\usepackage{comment}
\usepackage{multirow,array,amsmath}
\usepackage{enumitem}

\author{Ting Hua$^*$,
\enspace Yen-Chang Hsu$^*$,
  \enspace Felicity Wang,
  \enspace Qian Lou, 
  \enspace Yilin Shen, 
  \enspace Hongxia Jin\\
  Samsung Research America\\
  {\tt\{ting.hua,yenchang.hsu,f.wang1,qian.lou\}@samsung.com}\\
  {\tt\{yilin.shen,hongxia.jin\}@samsung.com}
  }

\title{Numerical Optimizations for Weighted Low-rank 
\\Estimation on Language Model} 

\usepackage{amsmath,amsfonts,bm}

\def\eqref#1{equation~\ref{#1}}

\def\1{\bm{1}}

\DeclareMathAlphabet{\mathsfit}{\encodingdefault}{\sfdefault}{m}{sl}
\SetMathAlphabet{\mathsfit}{bold}{\encodingdefault}{\sfdefault}{bx}{n}

\def\gL{{\mathcal{L}}}

\newcommand{\R}{\mathbb{R}}

\newcommand{\ie}{\textit{i}.\textit{e}.}
\newcommand{\eg}{\textit{e}.\textit{g}.}

\iffalse
    \newcommand{\yh}[1]{}
    \newcommand{\ting}[1]{}
    \newcommand{\yilin}[1]{}
    \newcommand{\needcite}[1]{}
\else
    \newcommand{\yh}[1]{{\color{green}{(Yen: #1)}}}
    \newcommand{\ting}[1]{{\color{orange}{(ting: #1)}}}
    \newcommand{\yilin}[1]{{\color{red}{(yilin: #1)}}}
    \newcommand{\needcite}{{\color{blue}{(cite)}}}
\fi

\begin{document}


\maketitle
  \def\thefootnote{*}\footnotetext{These authors contributed equally to this work.}\def\thefootnote{\arabic{footnote}}
\begin{abstract}
Singular value decomposition (SVD) is one of the most popular compression methods that approximate a target matrix with smaller matrices. However, standard SVD treats the parameters within the matrix with equal importance, which is a simple but unrealistic assumption. The parameters of a trained neural network model may affect the task performance unevenly, which suggests non-equal importance among the parameters.  Compared to SVD, the decomposition method aware of parameter importance is the more practical choice in  real cases. Unlike standard SVD, weighted value decomposition is a non-convex optimization problem that lacks a closed-form solution. 
We systematically investigated multiple optimization strategies to tackle the problem and examined our method by compressing Transformer-based language models.
Further, we designed a metric to predict when the SVD may introduce a significant performance drop, for which our method can be a rescue strategy.
The extensive evaluations demonstrate that our method can perform  better than current SOTA methods in compressing Transformer-based language models.
\end{abstract}

\section{Introduction}

Transformer-based language models  such as BERT \citep{devlin2018bert}
have obtained significant success in a variety of Natural Language Processing tasks, such as language modeling \citep{radford2018improving}, text classification \citep{wang2018glue}, question answering \citep{rajpurkar2016squad}, and summarization \citep{liu2019fine}. 
Despite
their success, these models usually contain millions or even billions of parameters, pre-trained by the large corpus. 
However, the downstream tasks may only focus on a specific scenario,
such that only a small amount of parameters in the big Transformer model will contribute to the performance of the target task.
Also, the massive size of Transformer models prohibits their deployments to resource-constrained devices. Therefore, compression of the Transformer-based language model attracts extensive interests.

Low-rank factorization \citep{golub1971singular,noach2020compressing} 
aims to approximate each parameter matrix in the trained model by two smaller matrices. 
This line of compression strategy will naturally inherit the knowledge of the big trained model without expensive generic re-training, and is the orthogonal direction to other compression approaches such as Knowledge distillation \citep{sun2019patient,sanh2019distilbert,jiao2019tinybert} or Quantization \citep{shen2020q,zhaoautomatic}.

However, applying standard SVD to approximate the learned weights often results in a significant task performance drop. 
Previous work shows that this phenomenon may be caused by a strong assumption held by the standard SVD, that the parameters in the matrix are equally crucial to the performance \citep{hsu2021language}.
Also, it has been observed that 
different parameters in Transformer models have different impacts on the overall task performance \citep{shen2020q}. 

Following FWSVD \citep{hsu2021language}, we utilize Fisher information \citep{Pascanu14revisitingnatural} to weigh the importance of parameters, so that the objective of matrix factorization will jointly consider matrix reconstruction error and the target task performance.
In the standard SVD, all the local minima are saddle points,  ensuring a closed-form global optimal solution \citep{srebro2003weighted}.   
This property no longer holds true to our new objective weighted by Fisher information. 
Without the closed-form solution, we revert to the numerical optimization methods to minimize the weighted objective. 
As our method can provide a more accurate solution than FWSVD \cite{hsu2021language}, we name our proposed method as TFWSVD (True Fisher Weighted SVD). 
Our results reveal the hybrid optimizer we called Adam\_SGD can best fit our problem, with its switching point estimated by the row-based analytic solution.
We also investigated the scenarios where SVD fails, under the guidance of the metric we introduced to measure the variance of parameter importance, with the example of analyzing the matrices within the Transformer blocks.

In summary, this work makes the following contributions: (1) we provide several optimization methods to search for the best numerical solution for low-rank estimation weighted by the Fisher information; (2) we perform extensive evaluations on various language tasks, showing our TFWSVD achieves better performance than the SOTA compression methods, and can further compress already compact models; (3) through the analysis of factorizing sub-structures inside the Transformer blocks, we provide the guide about when SVD may fail but TFWSVD can retain the performance.
\section{Background} 
\subsection{Model Compression with SVD}
Singular value decomposition (SVD) decomposes a matrix, \eg, $\bf{W} $$\in \R^{N \times M}$ into three matrices:
\begin{equation}
    \bf{W = US{V^T} \approx U_rS_r{V_r^T}},
\label{equ:svd}
\end{equation}
where $\bf{U}$$\in \R^{N \times l}$, $\bf{V}$$\in \R^{M \times l}$, and $l$ is the rank of matrix $\bf{W}$. $\bf{S}$ is a diagonal matrix of non-zero singular values $diag(\sigma_1, ,...,\sigma_l)$, where ${\sigma _1} \ge {\sigma _2} \ge  \cdots {\sigma _l} > 0$. 
$\bf{U_r}$, $\bf{S_r}$, and $\bf{V_r}$ represent the truncated matrices with rank $r$ and approximate the original matrix with a less total number of parameters.

The computation of a linear layer in neural networks can be rewritten as below with input data $\bf{X}$$\in\R^{1 \times N}$, weight matrix $\bf{W}$$\in \R^{N \times M}$, and bias $\bf{b}$$\in\R^{l\times M}$:
\begin{equation}
\bf{Z = XW + b  \approx (XU_rS_r){V_r^T} + b}.
\label{equ:linear_svd}
\end{equation}
The typical implementation of  factorization is to replace the large $\bf{W}$ with two smaller linear layers: 1) The weight matrix of the first layer is $\bf{US}$, which has $Nr$ parameters without bias. 2) While the weight matrix of the second layer is $\bf{V}$, with $Mr$ parameters plus bias. 
The truncation happens when $r$ is less than $l$. 
For example, if the total number of parameters for approximating $\bf{W}$ is $Nr+Mr$, then the reduced number of parameters will be $NM-(Nr+Mr)$.

\subsection{Fisher information}

A classical way to measure the importance of parameters is through the observed information, \ie Fisher information. It measures the amount of information that an observable dataset $D$ carries about a model parameter $w$. The accurate values of Fisher information are generally intractable since the computation will require marginalizing over the data $D$. 
In practice, the empirical Fisher information is estimated as follows:
\begin{equation}
\begin{aligned}
    I_w &=E\left[\left(\frac{\partial}{{\partial w }} \log p(\bf{D}|w )\right)^2\right]\\ & \approx \frac{1}{|\bf{D}|}\sum\limits_{i = 1}^{|\bf{D}|} {\left( {\frac{\partial }{{\partial w }}\gL(d_i;w)} \right)^2}={\hat I_w}.
\end{aligned}
    \label{equ:empirical_fisher}
\end{equation}
Given a target task objective $\gL$ (\eg, cross-entropy for a classification task), the estimated information $\hat I_w$ accumulates the squared gradients over the training data $d_i \in \mathcal{D}$. The parameters that cause large absolute gradient of the task objective will have a large value in $\hat I_w$, and are considered important to the target task.

\subsection{Related works}

The report of applying SVD to the Transformer layers is scarce. Several previous works applied SVD to compress the word embedding layer \citep{chen2018groupreduce, acharya2019online}.  Although \cite{noach2020compressing} combined knowledge distillation to fine-tune the resulting compressed model, they didn't address the issue of poor performance when fine-tuning is not applied. Experiments show that our proposed method can retrain most of the performance, providing a much better initialization for the fine-tuning. 

The use of Fisher information has appeared in many problem settings that also need to estimate the importance of model parameters, for example, to avoid catastrophic forgetting in continual learning \cite{kirkpatrick2017overcoming,hua2021hyperparameter} or model pruning
\cite{liu2021group,molchanov2019importance}. However, none of these work has explored its potential in assisting low-rank approximation for model compression.

Most previous work seeking the numerical solution for low-rank approximation is designed for unweighted cases,  with applications such as predicting the missing values recommendation system \cite{yu2014parallel,zhou2008large}.
Also, a few attempts have been made to solve the weighted low-rank approximation problem through EM-based algorithm \cite{srebro2003weighted}, or alternating least squares \citep{he2016fast}.

The closest previous work to this paper is FWSVD \citep{hsu2021language}, which points out that the ``even importance'' assumption held by SVD may cause a performance drop. FWSVD also utilizes Fisher information to weigh the importance of parameters. 
However, during the decomposition process, FWSVD assumes that parameters within each weight matrix row share the same importance value, which is still a strong assumption. Experimental results show that our TFWSVD can find more accurate solutions than FWSVD, as each parameter is associated with its own importance in TFWSVD. 

\section{Method}

\subsection{Low-rank factorization objective weighted by Fisher information}
The objective of the generic low-rank approximation is to minimize the Frobenius norm $||\bf{W-AB}||_2$, which is the sum squared differences of a reconstructed matrix $\bf{AB}$ to the target matrix $\bf{W}$.
As mentioned above, Singular value decomposition (SVD) can solve this problem efficiently by having $\bf{A=US}$ and $\bf{B=V^T}$. 
As the importance of each element $w_{ij}$ in $\bf{W}$ can be calculated through its Fisher information, we would like to find the reconstructed matrix  $\bf{AB}$ that minimizes the weighted Frobenius distance $J(\bf{A},\bf{B})$ as follows ($\otimes$ denotes element-wise multiplication):
\begin{equation}
\begin{aligned}
    J(\bf{A},\bf{B}) &=\bf{\hat I}\otimes (\bf{W}-\bf{AB})^2\\
    &=\sum\limits_{i,j} {{{\hat I}_{{w_{i,j}}}}} {({w_{i,j}} - \mathbf{a}_i^T\mathbf{b}_j)^2}.
    \label{equ:obj_no_reg}
\end{aligned}
\end{equation}
To prevent over fitting,  $L_2$ regularization terms controlled by  parameter $\lambda$ can be added to the objective, so that Equation (\ref{equ:obj_no_reg}) can be rewritten as: 

\begin{equation}
\begin{aligned}
    J(\bf{A},\bf{B}) &=\sum\limits_{i,j} {{{\hat I}_{{w_{i,j}}}}} {({w_{i,j}} - \mathbf{a}_i^T\mathbf{b}_j)^2} \\&+ \lambda (\sum\limits_i {||\mathbf{a}_i|{|^2}}  + \sum\limits_j {||\mathbf{b}_j|{|^2}} ).
    \label{equ:obj}
\end{aligned}
\end{equation}

\subsection{Optimization methods}
\label{sec:method:opt}
 SVD has an analytic solution, since  all of its  local minima are global.
However, this can not hold true when weights are introduced.
Without a closed-form solution, we discuss several numerical optimization methods to minimize $J(\bf{A},\bf{B})$.

\subsubsection{Alternating Least Squares}
Although the optimization problems in (\ref{equ:obj_no_reg}) and (\ref{equ:obj}) are non-convex,
they can be converted to quadratic problems with globally optimal solutions, if $\bf{A}$ or $\bf{B}$ is fixed.
Therefore, Alternating Least Squares (ALS) is suitable to solve such problems \citep{hastie2015matrix}.
ALS will alternately optimize $\bf{A}$ or $\bf{B}$ by keeping the other one fixed, and decrease $J(\bf{A},\bf{B})$ until convergence.
When the other matrix is fixed, minimizing $J(\bf{A},\bf{B})$ with respect to $\bf{A}$ or $\bf{B}$  is equivalent to minimize the following objectives:
\begin{equation}
\begin{aligned}
J({{\mathbf{a}}_{\mathbf{i}}}) &= ||{{{\mathbf{\hat I}}}_{{{\mathbf{W}}_{{\mathbf{[i,:]}}}}}}({{\mathbf{W}}_{{\mathbf{[i,:]}}}} - {\mathbf{B}}{{\mathbf{a}}_{\mathbf{i}}})|{|^2} + \lambda ||{{\mathbf{a}}_{\mathbf{i}}}|{|^2} \\
J({{\bf{b}}_{\bf{j}}}) &= ||{{{\bf{\hat I}}}_{{{\bf{W}}_{{\bf{[:,j]}}}}}}({{\bf{W}}_{{\bf{[:,j]}}}} - {\bf{A}}{{\bf{b}}_{\bf{j}}})|{|^2} + \lambda ||{{\bf{b}}_{\bf{j}}}|{|^2},
\end{aligned}
\end{equation}
which can lead to the closed-form solutions:
\begin{equation}
\begin{aligned}
{{\bf{a}}_{\bf{i}}}&={({{\bf{B}}^{\bf{T}}}{{{\bf{\hat I}}}_{{{\bf{W}}_{{\bf{[i,:]}}}}}}{\bf{B}} + \lambda {\bf{\Sigma}})^{{\bf{ - 1}}}}{{\bf{B}}^{\bf{T}}}{{{\bf{\hat I}}}_{{{\bf{W}}_{{\bf{[i,:]}}}}}}{{\bf{W}}_{{\bf{[i,:]}}}}\\
{{\bf{b}}_{\bf{j}}}&= {({{\bf{A}}^{\bf{T}}}{{{\bf{\hat I}}}_{{{\bf{W}}_{{\bf{[:,j]}}}}}}{\bf{A}} + \lambda {\bf{\Sigma}})^{{\bf{ - 1}}}}{{\bf{A}}^{\bf{T}}}{{{\bf{\hat I}}}_{{{\bf{W}}_{{\bf{[:,j]}}}}}}{{\bf{W}}_{{\bf{[:,j]}}}}
\end{aligned},
\end{equation}
where $\bf{\Sigma}$ is the identity matrix, while 
${{{\bf{\hat I}}}_{{{\bf{W}}_{{\bf{[i,:]}}}}}}$ and ${{{\bf{\hat I}}}_{{{\bf{W}}_{{\bf{[:,j]}}}}}}$ are the Fisher information vector of $i$-th row and $j$-th column in original matrix $\bf{W}$, respectively.

\subsubsection{Stochastic Gradient Descent}
Stochastic Gradient Descent (SGD) is
also shown to be effective for matrix factorization problems \cite{koren2009matrix}. 
Specifically in our problem, each update of SGD can be represented as:
\begin{equation}
\begin{aligned}
{{\bf{a}}_i} &\leftarrow {{\bf{a}}_i} + 2\eta ({e_{{w_{ij}}}}{{\bf{b}}_{\bf{j}}} - \lambda {{\bf{a}}_i}) \\
{{\bf{b}}_{\bf{j}}} &\leftarrow {{\bf{b}}_{\bf{j}}} + 2\eta ({e_{{w_{ij}}}}{{\bf{a}}_i} - \lambda {{\bf{b}}_{\bf{j}}}),
\end{aligned}
\end{equation}
where $\eta$ is the learning rate, and $ {e_{{w_{i,j}}}} = {{\hat I}_{{w_{i,j}}}}({w_{i,j}} - {\bf{a}}_{\bf{i}}^{\bf{T}}{{\bf{b}}_{\bf{j}}})$.
More generally, the iterations of SGD can be described as:
\begin{equation}
{\bf{h^{(k)}}} \leftarrow {\bf{h^{(k - 1)}}} - {\eta}\nabla J(\bf{h^{(k - 1)}}),
\label{equ:sgd_uniform}
\end{equation}
where $\bf{h}^{(k)}$ denotes the $k$-th iterate that can be substituted by $\bf{a}_i$ or $\bf{b}_j$.

\subsubsection{Adaptive Moment Estimation}

SGD will scale gradient uniformly in all directions, making the training process inefficient and sensitive to the learning rate. 
 Several adaptive methods have been proposed to overcome this shortcoming, among which Adaptive Moment Estimation (Adam) is one of the most widely used approaches \cite{kingma2015adam}.
Following the form of SGD updates shown in  (\ref{equ:sgd_uniform}), the Adam update iterations can be written as:
\begin{equation}
    \bf{h^{(k)}} \&\leftarrow {h^{(k - 1)}} - {\eta ^{(k - 1)}} \cdot \frac{{\sqrt {1 - \beta _2^{(k)}} }}{{1 - \beta _1^{(k)}}} \cdot \frac{{{m^{(k - 1)}}}}{{\sqrt {{v^{(k - 1)}} + \varepsilon } }},
\end{equation}
where $\bf{h}^{(k)}$ and $\eta$ are the same as Equation (\ref{equ:sgd_uniform}), ${\bf{m}^{(k - 1)}}$ and ${\bf{v}^{(k - 1)}}$ are calculated as follows:
\begin{equation}
    \begin{aligned}
{\bf{m}^{(k - 1)}} &= {\beta _1}{\bf{m}^{(k - 2)}} + (1 - {\beta _1})\nabla J(\bf{h^{(k - 1)}}) \\
{\bf{v}^{(k - 1)}} &= {\beta _2}{\bf{v}^{(k - 2)}} + (1 - {\beta _2})\nabla J{(\bf{h^{(k - 1)}})^2}.
    \end{aligned}
\end{equation}

Although Adam requires minimal tuning and enjoys fast initial progress, it is not without faults. 
Recent work has shown that the solutions found by Adam can be much worse at generalization than those found by SGD \cite{akiba2017extremely,ida2020improving}.

\subsubsection{Adam Switching to SGD}
Previous studies show that switching from Adam to SGD may contribute to the performance, however,
the switching point is crucial for the overall performance and usually is task-dependent  \cite{ida2020improving}.
Here we propose a simple method to calculate the switching point for our Fisher information weighted matrix factorization problem. 

Although weighted SVD does not have a closed-form solution when each element has its weight, the optimization problem (\ref{equ:obj}) has a close form in the case that elements within the same row share the same weight \citep{hsu2021language}. 
Therefore, we can calculate an approximate solution for the optimization problem (\ref{equ:obj}) based on row-wise Fisher information, which can be solved as the ``threshold'' for our switching point from Adam to SGD \citep{hsu2021language}.
If we define the importance for the row $i$ to be the summation of the row, \ie, $\hat {I}_{W_i}=\sum\limits_{j} \hat {I}_{W_{ij}}$ and diagonal matrix $\hat I= diag(\sqrt{\hat I_{W_1}},...,\sqrt{\hat I_{W_N}})$, then the optimization problem of Equation (\ref{equ:obj_no_reg}) can be written as:
\begin{equation}
J(\bf{A,B})  \approx \hat J({\mathbf{A},\mathbf{B}})  =||\hat IW - \hat IAB|{|_2}.
\label{equ:weight_svd_diag}
\end{equation}
Optimization problem (\ref{equ:weight_svd_diag}) can be solved by the standard SVD on $\hat IW$. 
If we denote $svd(\hat IW)=(U^*, S^*, V^*)$, then the solution of Equation (\ref{equ:weight_svd_diag}) will be $A={\hat I}^{ - 1} U^* S^*$, and $B=V^{*T}$. 
The value of $\hat J({\mathbf{A},\mathbf{B}}) $ is served as our switching point from Adam to SGD, that the training process will be optimized by Adam when the current loss is larger than $\hat J({\mathbf{A},\mathbf{B}})$, and then taken over by SGD when its loss is smaller than $\hat J({\mathbf{A},\mathbf{B}})$.

Besides the hard threshold calculated in (\ref{equ:weight_svd_diag}), we also set a soft threshold that restricts our unweighted reconstruction error with the same order of magnitude as that of SVD. Experiments in Section \ref{sec:exp:opt} show that our switching point can well balance the speed and convergence of the optimization process.

\subsection{Metric measuring when SVD may fail}
\label{sec:method:variance}
Besides an accurate solution to the $J(\bf{A},\bf{B})$,
whether TFWSVD can obtain a performance gain is also decided by the properties of the target matrix $\bf{W}$ itself. 
TFWSVD is to capture the different importance of parameters.
However, if the parameters in $\bf{W}$  equally contributed to the model performance,
then the standard SVD should be good enough. 
Driven by these factors, we are interested in this question: Is there a method that can ``foresee'' when SVD will fail, and TFWSVD can help  retain performance? 

Given target matrix $\bf{W}$, here we propose a simple but effective metric called Fisher information variance $\varphi ({\bf{W}})$, which is calculated as the variance of the $L_p$ normalization of its corresponding Fisher information ${{{{\bf{\hat I}}}_{\bf{W}}}}$:  
\begin{equation}
    \varphi ({\bf{W}}) = Var(\frac{{{{{\bf{\hat I}}}_{\bf{W}}}}}{{\max (||{{{\bf{\hat I}}}_{\bf{W}}}|{|_p},\varepsilon )}}).
    \label{equ:var}
\end{equation}
As shown in Section \ref{exp:transformer_block}, this metric can qualitatively measure whether the targeted matrix is too challenging to SVD and therefore needs help from TFWSVD.

\section{Experiment}

\begin{table*}
\centering
\caption{Results of CoNLL and GLUE benchmark. G-Avg means the average of GLUE tasks, A-Avg denotes the average of all tasks, including CoNLL.
Our method  is the best performer in terms of both average scores. 
}
\small
\label{tab:GLUE:sota}
\renewcommand\tabcolsep{3pt}
\resizebox{0.9\textwidth}{!}{
\begin{tabular}
{l|l|c|ccccccc|c|c}
\toprule
Task & \#Param & CoNLL & CoLA & MNLI & MRPC & QNLI & QQP & SST-2 & STSB & G-Avg & A-Avg \\\toprule
Bert\_base &109.5M &94.1 &56.2 &84.7 &87.4 &91.3 &87.8 &93 &88.5 &84.1 &85.4 \\\midrule
distilBERT &67.0M &93.2 &49.8 &82.2 &88.7 &89.3 &86.7 &90.4 &86.1 &81.9 &83.3 \\
MiniLMv2 &67.0M &92.2 &43.3 &84.0 &89.1 &90.6 &86.7 &91.4 &88.1 &81.9 &83.2 \\
TinyBERT6 &67.0M &93.2 &41.2 &83.9 &90.6 &90.6 &87.0 &92.1 &89.4 &82.1 &83.5 \\\midrule
SVD &66.5M &12.0 &2.7 &35.6 &61.4 &37.2 &60.0 &76.7 &26.8 &42.9 &39.0 \\
+ fine-tuning &66.5M &92.4 &40.5 &82.8 &84.1 &89.6 &87.3 &90.9 &85.7 &80.1 &81.6 \\\midrule
TVD &66.5M &51.6 &2.1 &58.8 &81.6 &66.9 &74.1 &83.1 &75.3 &63.1 &59.7 \\
+ fine-tuning &66.5M &93.4 &41.1 &82.4 &87.8 &89.2 &84.6 &90.3 &87.2 &80.4 &81.2 \\\midrule
FWSVD &66.5M &49.6 &13.5 &52.8 &81.2 &52.2 &65.7 &82.1 &68.6 &59.4 &58.2 \\
+ fine-tuning &66.5M &93.2 &49.4 &83.0 &88.0 &89.5 &87.6 &91.2 &87.0 &82.2 &83.6 \\\midrule
TFWSVD &66.5M &86.3 &20.6 &70.7 &79.6 &64.4 &76.0 &87.7 &69.0 &66.9 &69.3 \\
+ fine-tuning &66.5M &94.2 &52.2 &83.4 &89.0 &90.3 &86.9 &91.1 &88.5 &\textbf{83.1} &\textbf{84.4} \\
\bottomrule
\end{tabular}
}
\end{table*}

\subsection{Language tasks and datasets}

We evaluate our proposed methods and baselines on the General Language
Understanding Evaluation (GLUE) benchmark \citep{wang2019glue} and a token classification task.
More details about datasets and tasks can be found in Appendix \ref{sec:app:datasets}.


\subsection{Implementation details and baselines}
\label{sec:exp:baselines}
For generic compact methods (MiniLM, DistilBERT, and TinyBERT), we use the models provided by the original authors as the initialization, then directly fine-tune them on the training data of the target task. The fine-tuning is optimized by Adam with a learning rate of $2 \times 10^{-5}$ and batch size of 32 on one GPU.

Besides FWSVD \cite{hsu2021language} and our proposed TFWSVD, we also provide a baseline using first-order Taylor expansion for value decomposition (TVD). The details of TVD can be found in Appendix \ref{sec:app:TVD}. 

For  low-rank factorization methods (TFWSVD, FWSVD, TVD, and SVD),  we use the pre-trained 12-layer BERT model \citep{devlin2018bert} as the start.
And then, the large BERT model is fine-tuned on the task-specific data. 
Next, we apply the low-rank factorization, followed by another fine-tuning. 
We reported the results with and without fine-tuning to reveal the native results of low-rank factorization.

To make a fair comparison, only the linear layers in the transformer blocks are compressed in this work. The non-Transformer modules, such as the token embedding, are not compressed. Previous works \citep{chen2018groupreduce} have shown significant success in applying low-rank factorization to compress the embedding layer, which occupies 23.4M (21.3\%) parameters in the standard BERT model. Thus, the results we reported in this paper can be further improved by applying our method to non-transformer modules. 


All of our implements are created on the base of HuggingFace Transformer library \cite{wolf-etal-2020-transformers}.
The settings not  mentioned use the default configuration of the HuggingFace Transformer library.
We directly reported the results on the dev set for all datasets, as hyper-parameter searching is not involved in our experimental evaluations. 

\begin{table*}[htbp]
\centering
\caption{Results of CoNLL and GLUE benchmark with high compression rates. 
Compared to Table \ref{tab:GLUE:sota}, the advantages of TFWSVD over other two low-rank estimation methods are enlarged in the high compression rate settings.  
}
\small
\label{tab:GLUE:compact}
\renewcommand\tabcolsep{3pt}
\resizebox{0.9\textwidth}{!}{
\begin{tabular}
{l|l|c|ccccccc|c|c}
\toprule
Task & \#Param & CoNLL & CoLA & MNLI & MRPC & QNLI & QQP & SST-2 & STSB & G-Avg & A-Avg \\\toprule
Bert\_base &109.5M &94.1 &56.2 &84.7 &87.4 &91.3 &87.8 &93 &88.5 &84.1 &85.4 \\\midrule
SVD &49.9M &2.7 &0.2 &37.0 &0.0 &49.5 &36.9 &58.3 &16.5 &28.3 &25.1 \\
+ fine-tuning &49.9M &92.8 &19.3 &81.0 &82.0 &86.6 &86.9 &89.2 &80.6 &75.1 &77.3 \\
FWSVD &49.9M &6.0 &2.4 &38.1 &0.0 &49.5 &37.7 &58.4 &27.1 &30.4 &27.4 \\
+ fine-tuning &49.9M &92.9 &38.7 &81.4 &80.3 &88.0 &87.2 &88.4 &82.9 &78.1 &80.0 \\
TFWSVD &49.9M &6.0 &57.0 &55.3 &30.3 &49.6 &60.8 &79.1 &53.7 &55.1 &49.0 \\
+ fine-tuning &49.9M &93.5 &39.3 &82.2 &88.3 &88.8 &87.0 &89.9 &87.0 &\textbf{80.4} &\textbf{82.0} \\\midrule
SVD &37.2M &2.2 &0.0 &32.5 &0.0 &49.5 &3.4 &51.4 &5.5 &20.3 &18.1 \\
+ fine-tuning &37.2M &90.4 &13.8 &78.0 &82.0 &79.6 &84.1 &87.5 &58.7 &69.1 &71.7 \\
FWSVD &37.2M &0.0 &0.0 &35.4 &0.0 &49.5 &0.0 &51.0 &7.9 &20.5 &18.0 \\
+ fine-tuning &37.2M &3.5 &18.7 &78.2 &78.6 &82.3 &84.5 &88.9 &67.9 &71.3 &62.8 \\
TFWSVD &37.2M &11.6 &4.5 &35.8 &0.0 &49.5 &55.1 &72.7 &32.8 &35.8 &32.7 \\
w fine-tuning &37.2M &91.9 &21.4 &79.1 &85.0 &84.3 &85.9 &89.0 &86.0 &\textbf{75.8} &\textbf{77.8} \\
\bottomrule
\end{tabular}
}
\end{table*}

\begin{table*}[!tbp]
\centering
\caption{ Weighted error and standard error of different methods at their final stages. The weighted error is  $J(\bf{A},\bf{B})$ in (\ref{equ:obj_no_reg}), and the standard error is $||\bf{W}-\bf{A}\bf{B}||_2$.  Adam and Adam\_SGD are trained 50,000 steps, while ALS is trained for 2.5 million steps and SGD is trained for 3 million steps.}
\small
\label{tab:opt}
\resizebox{0.9\textwidth}{!}{
\begin{tabular}{c|cc|cccc}
\toprule
& SVD & Closed-Form & SGD & Adam & ALS & Adam\_SGD \\
\midrule
Weighted error & 1.34E-05 & 9.87E-06 & 6.71E-06 & 1.06E-05 & 9.30E-06 & 1.28E-06 \\
Standard error & 6.57E-11 & 6.80E-11 & 2.38E-10 & 4.43E-11 & 2.31E-07 & 6.14E-11 \\
\bottomrule
\end{tabular}
}
\end{table*}

\subsection{Performance comparisons with SOTA}
\label{sec:exp:overall}
Table \ref{tab:GLUE:sota} reports the results of GLUE tasks and one NER task CoNLL. 
Our TFWSVD with 66.5M parameters obtains G-Avg score of $83.1$ and A-Avg score of $84.4$, which are better than the scores of SOTA models (MiniLMv2, TinyBERT6, distilBERT) requiring generic re-training. 
TFWSVD consistently yields good results on all the tasks, while the other generic re-training methods display obvious performance variance among different tasks. 
For example, TinyBERT6 is good at the STSB task but poor at CoLA; oppositely, distilBERT has strong performance on CoLA but is weak at STSB.

In the comparisons among low-rank factorization methods (TFWSVD, FWSVD, TVD, and SVD),
our TFWSVD beats other methods with apparent better performance in both scenarios with or without fine-tuning. 
One interesting phenomenon is that TVD can yield better results than SVD without fine-tuning. However, after fine-tuning, its advantages disappear, and SVD can achieve better average scores (G-Avg and A-Avg). 
This is not surprising.
Similar to our proposed TFWSVD, TVD is also a loss-aware method that definitely will be better than the loss-unaware SVD. 
But this gap can be narrowed or even eliminated with fine-tuning since SVD can also ``see'' the loss in this case. 
Therefore, within loss-aware methods, the weighting metric itself plays an important role in keeping the performance advantage. 
Also, TFWSVD obtains better performance than FWSVD, which indicates it is too ``aggressive'' for FWSVD to assume that parameters in the same row share the same importance.  

\subsection{Under high compression rates}
\label{sec:exp:low_compression_rate}

In this part, we compared low-rank methods under high compression rates. Because TVD didn't show an apparent advantage over SVD, here we mainly focus on comparing our proposed TFWSVD, FWSVD, and standard SVD.

As can be seen from Table \ref{tab:GLUE:compact}, TFWSVD always enjoys obvious advantages over the other two methods. 
Also, the performance gap between TFWSVD and FWSVD is enlarged as the compression rate goes higher.
In fact, under the extremely compact setting of 37.2M, FWSVD shows worse performance compared to SVD. 
This phenomenon further proves that the row-based importance assumption held by FWSVD may hurt the performance. 
While the privilege of TFWSVD always exists and becomes more prominent in the high compression rate of 49.9M and 37.2M. 
Especially in the scenario without fine-tuning, which can best reveal the pure performance of low-rank factorization, TFWSVD has performance scores almost double that of FWSVD and SVD.  

\begin{figure*} [htbp]
  \begin{subfigure}[b]{.5\textwidth}
  \raggedleft
   \includegraphics[height=4.9cm]{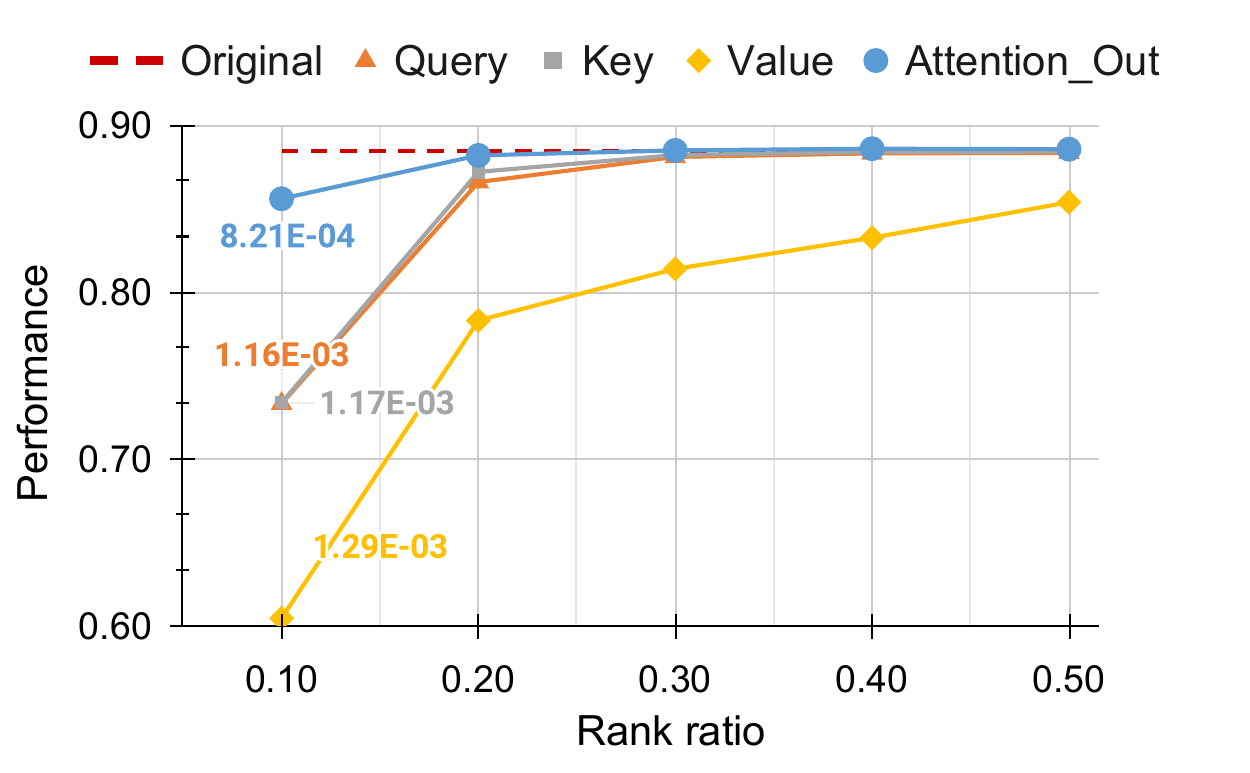}
       \caption{SVD performance on  768 $\times$ 768 dimension matrix
       }
       \label{fig:svd-structure:qkv-ao}
       \end{subfigure}
  \begin{subfigure}[b]{.5\textwidth}
   \centering
   \includegraphics[height=4.9cm]{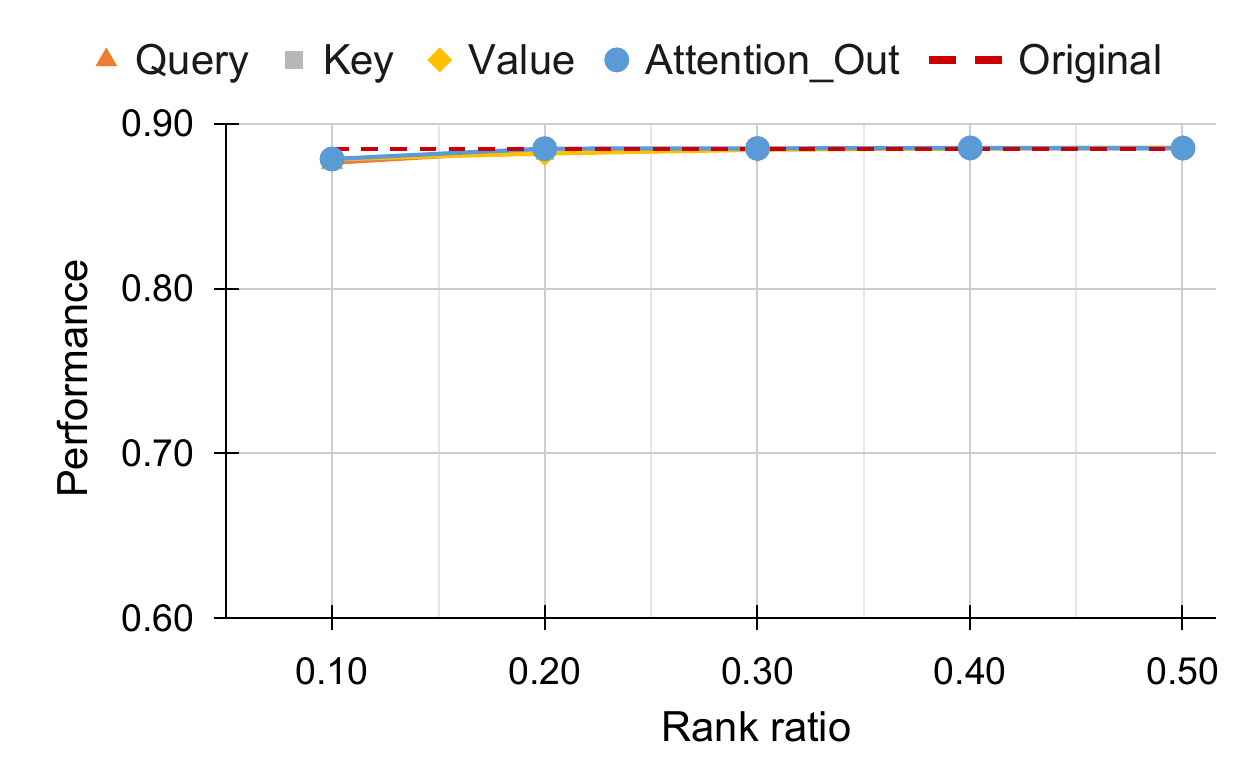}
       \caption{TFWSVD performance on  768 $\times$ 768 dimension matrix
       }
       \label{fig:fvd-structure:qkv-ao}
       \end{subfigure}
\caption{The performance of SVD and TFWSVD on the STSB task, when only factorizing a particular type of  sub-structures (Key, Query, Value, Attention) in Transformer blocks.
The red dash line denotes the original performance. 
The numbers marked besides the lines are the metric $\varphi ({\bf{W}})$ calculated by Equation (\ref{equ:var}).
The values of $\varphi ({\bf{W}})$ can well predict the performance of SVD, that matrix with a larger $\varphi ({\bf{W}})$ will always end up with a larger performance drop after applying SVD.
}
\label{fig:svd-fvd-structure}
\end{figure*}


\subsection{Optimization methods}
\label{sec:exp:opt}

In this part, we compare optimization procedures mentioned in Section \ref{sec:method:opt} to identify the best optimizer for our approximation problem.
\begin{figure}[htbp]
\centering
\includegraphics[width=\columnwidth]{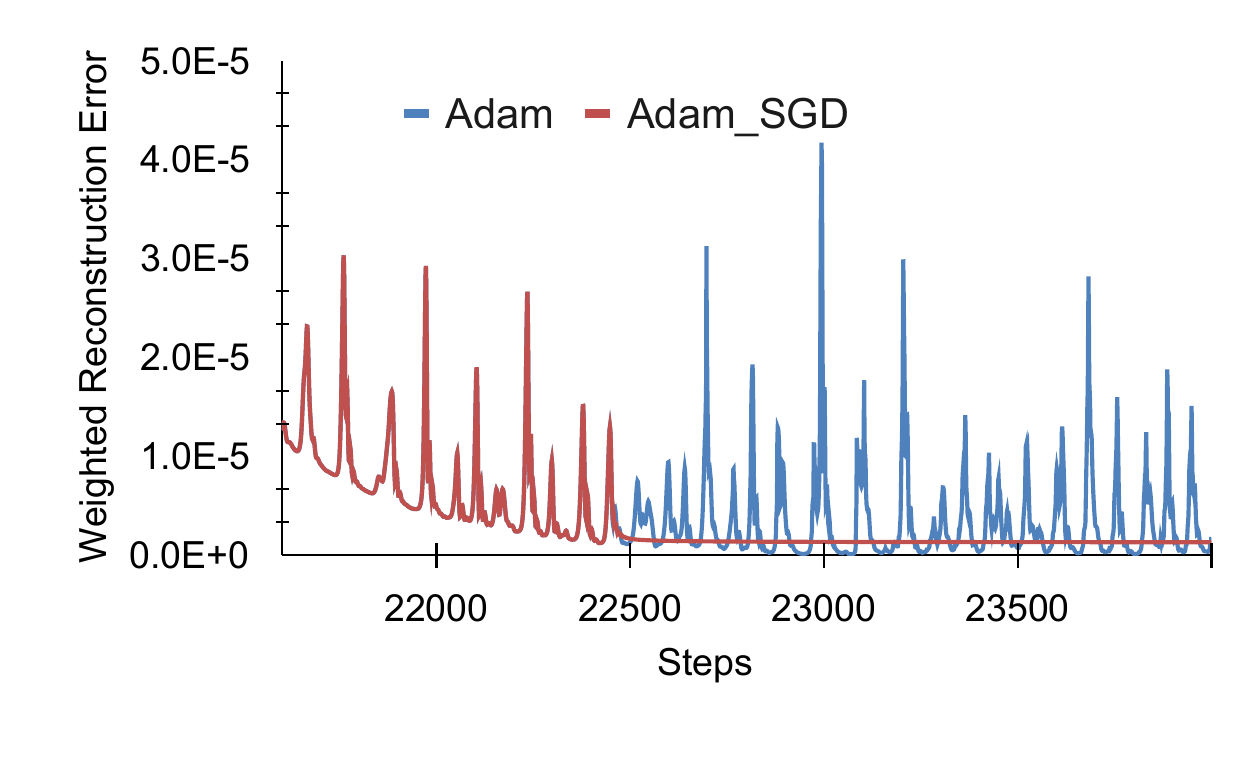}
\caption{Numerical experiments comparing Adam and Adam\_SGD on STSB. 
For Adam\_SGD, the switching point from Adam to SGD is around step 22400.
} 
\label{fig:adam_sgd}
\end{figure}

\subsubsection{ALS and SGD}
\label{sec:method:alsVSSGD}
In order to update the latent vectors, ALS needs $O(r^2)$ time to form the $r \times r$ matrix, with an additional $O(r^3)$ time to solve the least-squares problem.
Therefore, to reconstruct the target matrix $\bf{W} \in$ $\R^{N\times M}$ with rank $r$, the time complexity of one ALS iteration is $O((M+N)r^3+MNr^2)$.
It has been pointed out that ALS can be speeded up by parallelly updating each row of $\bf{A}$ or $\bf{B}$ independently \cite{zhou2008large}. 
While for SGD, the time complexity per iteration is only $O(MNK)$. 
Compared to ALS, SGD seems to be faster in terms of the time complexity for one iteration.
However, typically it requires more iterations than ALS to achieve relatively good performance \cite{yu2014parallel}.
As shown in Table \ref{tab:opt}, in order to obtain the performance close to that of Adam/Adam\_SGD, ALS and SGD need $50 \sim 60$ times more steps, which makes them impractical to be used in the real-world Transformer compression. 
Therefore, in the rest of this part, we will focus on comparing the performance of Adam and Adam\_SGD.

\subsubsection{Adam and Adam\_SGD}
The goal of hybrid optimizer Adam\_SGD is to combine the benefits of Adam (fast initial progress and minimal efforts in hyperparameter tuning) and SGD (good convergence and  generalization).

As seen from Figure \ref{fig:adam_sgd}, Adam and Adam\_SGD share the same trajectory in the initial steps.
After the switching point (around 22400 in Figure \ref{fig:adam_sgd}), Adam\_SGD converges to a low error solution (1.28E-06 as shown in Table \ref{tab:opt}), which is much smaller than the row-based analytic solution (9.87E-06).  
In contrast, Adam fluctuates in performance and ends with a much larger error 1.06E-05. 
These phenomena prove the effectiveness of Adam\_SGD in solving the weighted Frobenius distance optimization problem in (\ref{equ:obj_no_reg}) and (\ref{equ:obj}).
And the reconstruction errors of final solutions obtained by Adam\_SGD are 5$\sim$10 times smaller than the row-wise approximations.

\subsection{Fisher information variance}
\label{exp:transformer_block}
What is the secret behind TFWSVD's good performance on Transformer-based model compression?
In this part, we utilize the metric Fisher information variance $\varphi ({\bf{W}})$ introduced in Section \ref{sec:method:variance} to reveal the secret by analyzing the sub-structures inside the Transformer blocks. 

According to the implementation of HuggingFace Transformer library \cite{wolf-etal-2020-transformers}, there are five kinds of linear layers within the Transformer block, which can be set into two groups by their dimensions: Query, Key, Value, and Multi-head Attention layers are matrices with the dimension of $768 \times 768$; and two feed-forward layers called Intermediate and Output, are $768 \times 3072$ in dimension. 
Figure \ref{fig:svd-fvd-structure}   plot the performance changes along with varying the rank ratio for matrices with the dimension of $768 \times 768$,  when only decomposing one type of sub-structure. 
More results are plotted in Figure \ref{fig:svd-fvd-structure-3000} in Appendix.

Compared to the overall performance comparison in Section \ref{sec:exp:overall}, the purpose of this experiment is to evaluate the performance of SVD and TFWSVD on the finer-level sub-structures within Transformer blocks. Taking Figure \ref{fig:svd-structure:qkv-ao} for example, the yellow line denoting ``Value'' means: only the ``Value'' sub-structures are decomposed by SVD, while other types of  sub-structures are kept the same as the original model. 
We calculate the Fisher information variance $\varphi ({\bf{W}})$ via Equation (\ref{equ:var}), and mark the values besides the corresponding sub-structures. 
Several observations can be made from Figure \ref{fig:svd-fvd-structure}.

{\noindent}\textbf{Different matrix has different sensitiveness to SVD}. As shown in Figure \ref{fig:svd-structure:qkv-ao}, Attention\_out layer is relatively easy to compress. Even with standard SVD, it can still achieve good performance as low as a rank ratio 0.1. While compressing matrix Intermediate is rather difficult, its performance will drop down to 17\% with a rank ratio 0.1. 

{\noindent}\textbf{Metric Fisher information variance $\varphi ({\bf{W}})$ can `foresee'' the performance of SVD}. In Figure \ref{fig:svd-structure:qkv-ao}, decomposing sub-structures with larger $\varphi ({\bf{W}})$ via SVD will always cause the more serious performance drop. Especially, the performance changes of sub-structure Query and Key are almost identical, and their  $\varphi ({\bf{W}})$ are extremely close (1.16E-03 for Query and 1.17E-03 for Key). This phenomenon implies the metric $\varphi ({\bf{W}})$ can well reflect the variance of parameter importance within the matrix, and therefore can be a good performance indicator for SVD.

{\noindent}\textbf{TFWSVD can always help improve the performance}.
 Figure \ref{fig:fvd-structure:qkv-ao}  shows that applying TFWSVD will bring significant performance gain to all the sub-structures. Especially for the challenging matrix Intermediate (Figure \ref{fig:fvd-structure-3000} in Appendix), TFWSVD achieves an excellent performance of 60\% at a low-rank ratio 0.1, which is a 200\% improvement compared to the corresponding SVD performance 17\%.

\subsection{Compress the already compact models}

\begin{table}[]\centering
\caption{Results of further compressing the compact models. TFWSVD successfully reduces the size of the light-weight models, and achieves slightly better performances than the original compact models. 
}\label{tab:compact_compress}
\scriptsize
\renewcommand\tabcolsep{3.2pt}
\resizebox{0.49\textwidth}{!}{
\begin{tabular}{l|c|ccc|r}\toprule
&Param(M) &CoNLL &MRPC &STSB &Avg \\\midrule
DistillBERT &67.0 &94.0 &88.7 &86.1 &89.6 \\
w SVD &43.6 &93.2 &82.9 &83.0 &86.4 \\
w FWSVD &43.6 &93.4 &87.9 &84.1 &88.5 \\
w TFWSVD &43.6 &93.6 &89.0 &87.3 &\textbf{90.0} \\\midrule
MiniLMv2 &67.0 &93.2 &89.0 &88.1 &90.1 \\
w SVD &43.6 &93.2 &89.0 &86.4 &89.5 \\
w FWSVD &43.6 &93.3 &88.8 &87.9 &90.0 \\
w TFWSVD &43.6 &93.6 &90.0 &88.7 &\textbf{90.8} \\\midrule
TinyBERT6 &67.0 &93.2 &90.5 &89.4 &91.0 \\
w SVD &43.6 &93.0 &88.5 &88.3 &89.9 \\
w FWSVD &43.6 &93.1 &89.0 &88.7 &90.3 \\
w TFWSVD &43.6 &93.2 &90.7 &89.5 &\textbf{91.1} \\\midrule
TinyBERT4 &14.3 &85.6 &89.0 &85.9 &86.8 \\
w SVD &11.9 &88.9 &87.1 &84.7 &86.9 \\
w FWSVD &11.9 &88.5 &87.6 &86.1 &87.4 \\
w TFWSVD &11.9 &88.6 &88.8 &87.1 &\textbf{88.2} \\
\bottomrule
\end{tabular}
}
\end{table}
The matrix factorization direction is thought to be orthogonal to other compression methods such as knowledge distillation.
But in practice, performance drops are often observed when combining the different lines of compression technologies. 
Table \ref{tab:compact_compress} reports the results of applying TFWSVD, FWSVD, and SVD to compress the lightweight models further.
In general, TFWSVD can reduce 30\% more parameters for the compact models, with even improved performance.
In fact, the performance gains by applying TFWSVD are observed on all compact models in Table \ref{tab:compact_compress}, while both SVD and FWSVD will cause performance drops more or less when combined with those compact models. 
These results indicate that SVD and FWSVD may not be fully integrated with other compression technologies due to the 
the strong assumptions they held.
And our TFWSVD can best explore the potential of combining other lines of compression methods with matrix factorization.

\subsection{Discussion}
The incorrect predictions from the trained model will bring larger gradients than the correctly labeled examples, which means these incorrect predictions may be the better choices to compute Fisher information. It is different from our intuitions, but not surprising, since all these examples can reflect the features of trained parameters. In fact,  the mislabeled examples may better ``describe'' the features of the trained model (for example, these examples are around the boundary).
Also, we can use incorrect-only labels to estimate the Fisher information to further reduce computation time.
More details can be found in Appendix \ref{sec:app:labels}.

\section{Conclusion}
Unlike SVD, there is no closed-form solution for the weighted low-rank estimation problem, which therefore has to be approximated via numerical optimization methods. 
We managed to obtain the practical solutions through our hybrid Adam\_SGD optimizer with the specially designed switching point.
Our TFWSVD consistently works better than other low-rank factorization methods (FWSVD, TVD, and SVD).
Compared to SOTA methods that requiring expensive generic re-training, our TFWSVD shows more stable performance on various tasks.
Also, TFWSVD  can efficiently further compress and optimize the already compact models.
We also investigate the properties of the targeted matrix, where that SVD may fail, and TFWSVD can be the rescuer. 
We believe our TFWSVD could be the best alternative to SVD for language model compression.

\clearpage
\section{Limitations} 
The most significant limitation of TFWSVD is that it may cost more time than SVD and FWSVD. 
Compared to FWSVD, TFWSVD will need more time in numerical optimization, which is decided by the number of parameters in a model and is fixed for all downstream tasks. For GLUE tasks trained with the BERT model, the extra time cost of TFWSVD is around 1.5 V100 GPU hours. 
Also, both TFWSVD and FWSVD need time for Fisher information calculation. For example, this calculation takes about 8 minutes on SST-2 task. And it can be further reduced to around 5 seconds if we only use incorrect predictions (see details in Appendix \ref{sec:app:labels}).

In summary, compared to SVD and FWSVD, TFWSVD will cost at least 1.5 more V100 GPU  hours when compressing the BERT model for a GLUE task. 
This cost is worthy, considering the stable performance gain that TFWSVD can bring. 
On the other hand, TFWSVD  is still a much faster choice than the generic re-trained compact models such as distilBERT and MiniLM. For example, distilBERT needs 720 V100 GPU hours for re-training a BERT model. Similar to SVD and FWSVD, our TFWSVD can avoid such expensive re-training and can be applied to the directly downloaded BERT model.  

\section{Ethical Considerations} 
Our work is to better compress the language model with an improved low-rank estimation method. 
For our experiments, we used open datasets without sensitive information, which have been widely mentioned in previous work.
No license is required for the GLUE dataset, and we have purchased the license for the CoNLL dataset.
In the application of our model, we do not think there is an obvious issue that may lead to a risk to ethics.

\bibliography{ACL2022_main}
\bibliographystyle{acl_natbib}


\clearpage
\begin{table*}[htbp]
\caption{Performance comparison of using correct/incorrect labeled examples in the estimation of Fisher information.
All results here are without fine-tuning. \#Examples denotes the number of corresponding examples, I-AVG means the average importance score, F1 and ACC are task specific performance metrics.}
\label{tab:wrong-label}
\renewcommand\tabcolsep{2.0pt}
\resizebox{1\textwidth}{!}{
\begin{tabular}{l|ccc|ccc|ccc|ccc}
\hline
              & \multicolumn{3}{c|}{MNLI}                                                               & \multicolumn{3}{c|}{QNLI}                                                               & \multicolumn{3}{c|}{QQP}                                                               & \multicolumn{3}{c}{SST2}                                                               \\ \hline
              & \multicolumn{1}{l|}{\#Examples} & \multicolumn{1}{l|}{I-AVG} & \multicolumn{1}{l|}{ACC} & \multicolumn{1}{l|}{\#Examples} & \multicolumn{1}{l|}{I-AVG} & \multicolumn{1}{l|}{ACC} & \multicolumn{1}{l|}{\#Examples} & \multicolumn{1}{l|}{I-AVG} & \multicolumn{1}{l|}{F1} & \multicolumn{1}{l|}{\#Examples} & \multicolumn{1}{l|}{I-AVG} & \multicolumn{1}{l}{ACC} \\ \hline
Correct-only  & \multicolumn{1}{c|}{374345}     & \multicolumn{1}{c|}{1.21}  & 69.68                    & \multicolumn{1}{c|}{103114}     & \multicolumn{1}{c|}{0.27}  & 60.79                    & \multicolumn{1}{c|}{353888}     & \multicolumn{1}{c|}{0.69}  & 75.49                   & \multicolumn{1}{c|}{66567}      & \multicolumn{1}{c|}{0.05}  & 87.50                   \\
Incorect-only & \multicolumn{1}{c|}{18357}      & \multicolumn{1}{c|}{3.55}  & 70.63                    & \multicolumn{1}{c|}{1629}       & \multicolumn{1}{c|}{1.01}  & 63.48                    & \multicolumn{1}{c|}{9958}       & \multicolumn{1}{c|}{2.27}  & 75.95                   & \multicolumn{1}{c|}{782}        & \multicolumn{1}{c|}{0.29}  & 87.39                   \\
All           & \multicolumn{1}{c|}{392702}     & \multicolumn{1}{c|}{4.76}  & 70.65                    & \multicolumn{1}{c|}{104743}     & \multicolumn{1}{c|}{1.28}  & 64.42                    & \multicolumn{1}{c|}{363846}     & \multicolumn{1}{c|}{2.96}  & 76.00                   & \multicolumn{1}{c|}{67349}      & \multicolumn{1}{c|}{0.34}  & 87.73                   \\ \hline
\end{tabular}}
\end{table*}
\appendix
\begin{figure*} [htbp]
  \begin{subfigure}[b]{.49\textwidth}
  \raggedleft
\includegraphics[height=4.1cm]{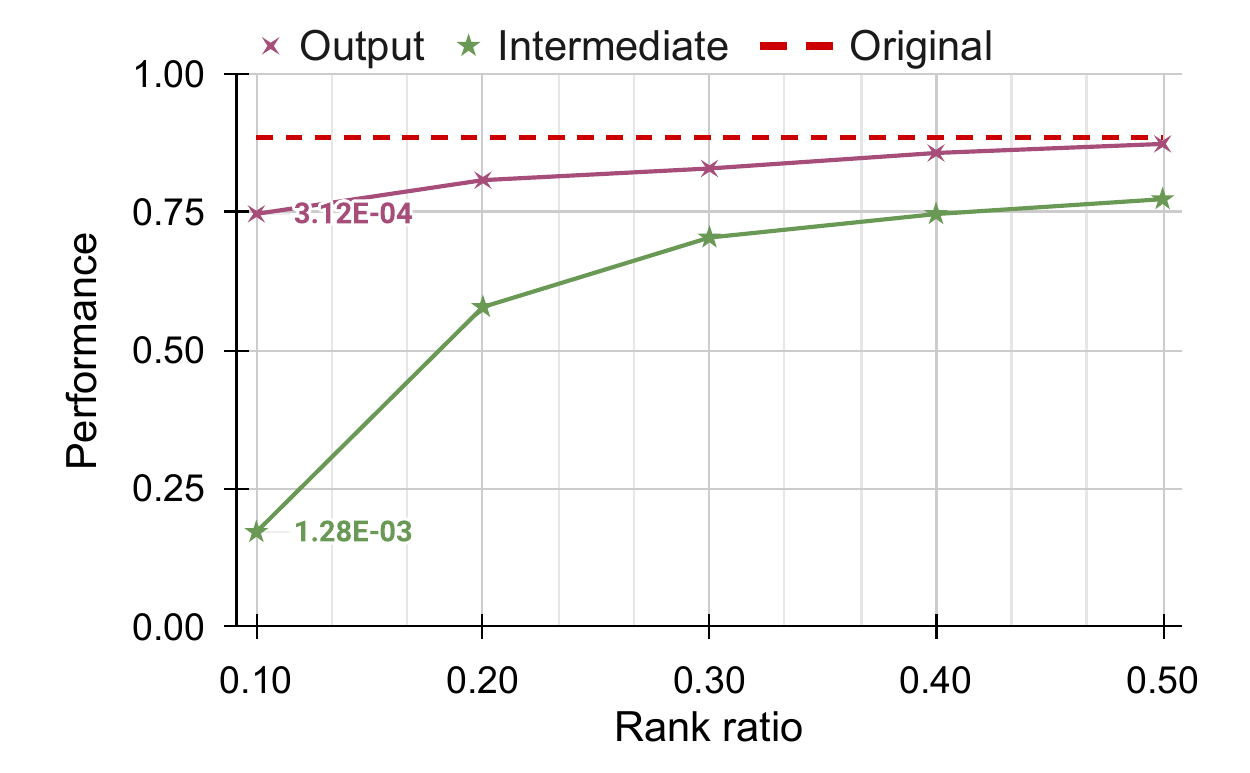}
       \caption{SVD performance on  3072 $\times$ 768 dimension matrix
       }
       \label{fig:svd-structure-3000}
       \end{subfigure}
  \begin{subfigure}[b]{.49\textwidth}
   \centering
\includegraphics[height=4.1cm]{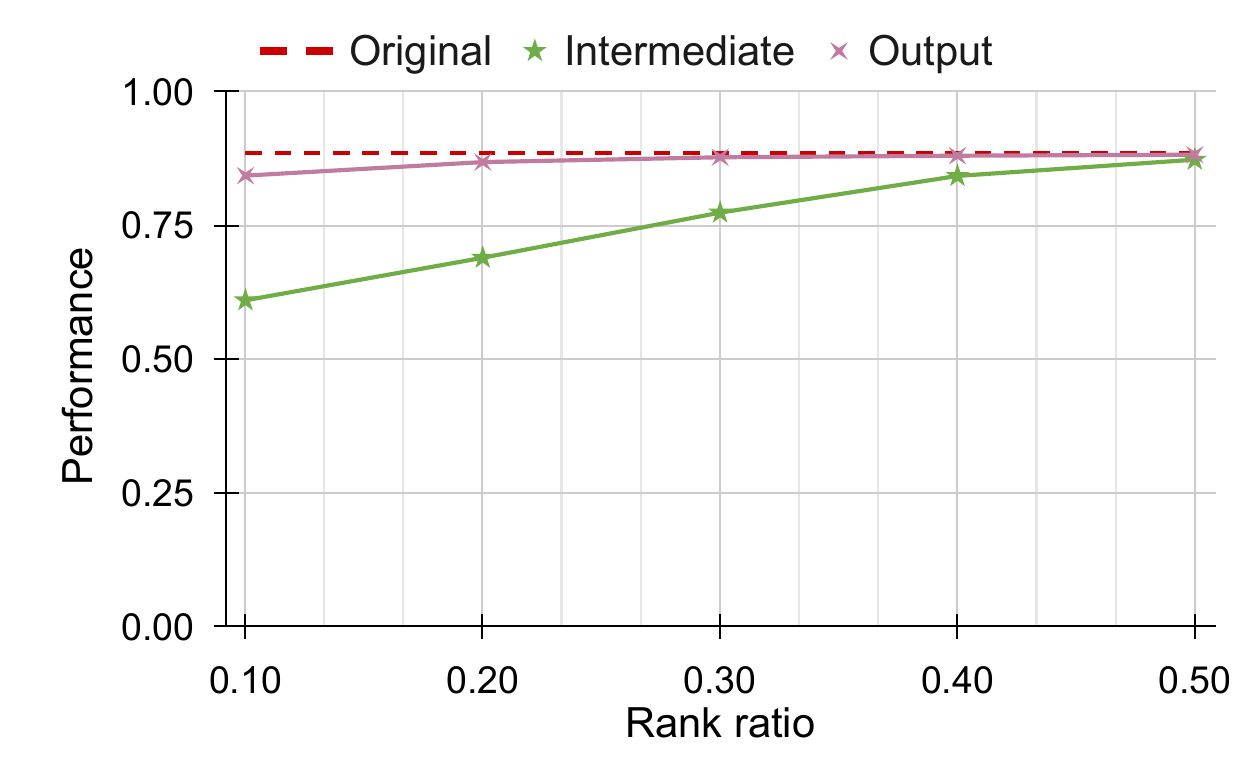}
       \caption{TFWSVD performance on  3072 $\times$ 768 dimension matrix
       }
       \label{fig:fvd-structure-3000}
       \end{subfigure}
\caption{The performance of SVD and TFWSVD on the STSB task, when only factorizing a particular type of  sub-structures (Intermediate, or Output) in Transformer blocks.
}
\label{fig:svd-fvd-structure-3000}
\end{figure*}
\section{Details of tasks and datasets}
\label{sec:app:datasets}

We include two single sentence tasks: CoLA \citep{warstadt2018neural} measured in Matthew's correlation, SST2 \citep{socher2013recursive} measured in classification accuracy; three sentence similarity tasks: MRPC \citep{dolan2005microsoft} measured in F-1 score, STS-B \citep{cer-etal-2017-semeval} measured in Pearson-Spearman correlation, QQP \citep{chen2018quora} measured in F-1 score; and three natural
language inference tasks: MNLI \citep{williams2018broad} measured in classification accuracy with the average of the matched and mismatched subsets, QNLI \citep{rajpurkar2016squad} measured in accuracy. The token classification task we used is the named entity recognition (NER) on the CoNLL-2003 dataset \citep{sang2003introduction}. In summary, our evaluation includes eight different natural language tasks.

\section{TVD}
\label{sec:app:TVD}
In this section, we provide the details about the baseline using first-order Taylor expansion for value decomposition (TVD). 
Following \cite{hou2020dynabert,voita2019analyzing}, we utilize the first-order Taylor expansion as the alternative importance score for matrices:
\begin{subequations}
\begin{align}
   T_w &=| \gL - \gL_{\neg w}| \label{TVD1}\\ 
   &=| \gL - (\gL - \frac{\partial\gL}{\partial w}(w-0)+R_{w=0})| \label{TVD2} \\
   &\approx |\frac{\partial\gL}{\partial w} w|. \label{TVD3}
\end{align}
\end{subequations}
As shown in \eqref{TVD1}, the intuition behind TVD is that the importance of a parameter $w$ can be calculated by the  variation in the training loss  when removing this parameter.
If we ignore the remainder $R_{w=0}$, then we can simply calculate the importance via \eqref{TVD3}, which is the product of the parameter value and its 1st-order gradient.

\section{Effect of incorrect predictions}
\label{sec:app:labels}

In this section, we evaluate whether the incorrect predictions will have negative impacts on the estimation of Fisher information.
In order to achieve this goal, we report the performance of classification tasks in Table \ref{tab:wrong-label}, when we use incorrectly/correctly predicted examples to estimate Fisher information.

Several observations can be made as follows.
 \textbf{First, the final performances are close, no matter using correct-only examples, incorrect-only examples, or all examples}. It demonstrates all kinds of examples can somehow reflect the importance of parameters. Meanwhile, the performances using all examples are always the best, confirming the better estimation of empirical Fisher information with more data.  
  \textbf{Second, using the incorrect-only examples will generate bigger values and better performance than using correct-only predictions}.
    Although only 1-2\% examples are incorrectly predicted, choosing these examples to estimate Fisher information will produce close numbers to those generated using all examples.  This is because Fisher information  is calculated via loss, and incorrect predictions will produce larger losses than the correct predictions.
    And compared to using correct-only examples, computations through incorrect-only examples may even bring better results for most tasks. 

In summary, the wrong labeled examples will generate larger Fisher information, but it doesn't mean that the Fisher information learned from the incorrectly labeled data is ``wrong''.
Instead, the mislabeled examples are better choices than correct predictions, which will produce better results with fewer computations.

\section{Training Time}
\label{sec:app:trainingTime}
This part discusses the training time of different approaches mentioned in this paper.

\noindent\textbf{TFWSVD versus FWSVD, TVD, and SVD}.
First, we compare the time costs of low-rank estimation methods SVD, TVD, FWSVD, and our proposed TFWSVD.
In general, SVD is the fastest method that can be done immediately as it has a close-form solution. 
FWSVD is the second fast method, which needs time for Fisher information calculation.
TFWSVD and TVD will cost more time in the numerical optimization process. 
\begin{enumerate}
    \item FWSVD versus SVD:  Compared to SVD, FWSVD needs extra time for Fisher information calculation. The time of this process is similar to one epoch of regular training.  For example, SST-2 task in this paper takes about 8 minutes to calculate the Fisher information.  This process is generally fast, and it can be further reduced to around 5 seconds if we only use incorrect predictions ( e.g., 1\% of all examples, mentioned in Appendix \ref{sec:app:labels}).
    \item TFWSVD versus FWSVD:  Compared to FWSVD, TFWSVD needs extra time for factorizing the weighted matrices through optimizations. The time cost of factorization is decided by the number of parameters in a model, and is fixed for all its downstream tasks. For GLUE tasks trained with the BERT model, TFWSVD will cost 1.5 more V100 GPU hours than FWSVD. 
    \item TFWSVD versus TVD: TFWSVD and TVD will cost the same time as these approaches are almost the same except for the weighting scheme. 

\end{enumerate}

\noindent\textbf{TFWSVD versus generic re-trained models}. The generic re-trained compact models such as distilBERT and MiniLMv2 require a large amount of re-training time. For example, distilBERT needs 720 V100 GPU hours for retraining a pre-trained BERT  model. Compared to these methods,  our TFWSVD is much faster, since TFWSVD can be applied to the directly downloaded BERT model without expensive re-training.

\clearpage

\clearpage

\end{document}